\documentclass[review]{elsarticle}

\usepackage{times}  \usepackage{helvet}  \usepackage{courier}  \usepackage{url}  
\usepackage{graphicx}
\usepackage[caption=false]{subfig}
\usepackage{epstopdf}
\usepackage{algorithm}
\usepackage{amsmath, amssymb}
\usepackage[noend]{algpseudocode}
\usepackage{caption} 
\journal{Journal of \LaTeX\ Templates}

\begin{document}

\begin{frontmatter}
\title{Intelligent Querying for Target Tracking in Camera Networks using Deep Q-Learning with n-Step Bootstrapping \tnoteref{mytitlenote}}
\tnotetext[mytitlenote]{Fully documented templates are available in the elsarticle package on \href{http://www.ctan.org/tex-archive/macros/latex/contrib/elsarticle}{CTAN}.}

\author{Anil Sharma, Saket Anand, Sanjit K. Kaul}
\address{Indraprastha Institute of Information Technology (IIIT-Delhi), \\ Delhi, India}




\begin{abstract}
Surveillance camera networks are a useful infrastructure for various visual analytics applications, where high-level inferences and predictions could be made based on target tracking across the network. Most multi-camera tracking works focus on target re-identification and trajectory association problems to track the target. However, since camera networks can generate enormous amount of video data, inefficient schemes for making re-identification or trajectory association queries can incur prohibitively large computational requirements. In this paper, we address the problem of intelligent scheduling of re-identification queries in a multi-camera tracking setting. To this end, we formulate the target tracking problem in a camera network as an MDP and learn a reinforcement learning based policy that selects a camera for making a re-identification query. The proposed approach to camera selection does not assume the knowledge of the camera network topology but the resulting policy implicitly learns it. We have also shown that such a policy can be learnt directly from data. Using the NLPR MCT and the Duke MTMC multi-camera multi-target tracking benchmarks, we empirically show that the proposed approach substantially reduces the number of frames queried. 
\end{abstract}

\begin{keyword}
Camera Networks\sep Deep Reinforcement Learning \sep Target Tracking \sep Multi-Camera Tracking
\MSC[2010] 00-01\sep  99-00
\end{keyword}

\end{frontmatter}


\section{Introduction}

Camera networks are becoming ubiquitous in smart cities where monitoring of urban environments has numerous applications like traffic management, law enforcement and security and automated surveillance. In these scenarios, camera sensors are deployed in public spaces like road intersections, common areas in residential, commercial and government complexes to collect data, which is transmitted, stored and analysed by the government or local authorities. For example, surveillance cameras in residential and commercial complexes can be used to identify and track trespassers and unauthorized personnel or for forensic analysis for investigating crimes.  

For these applications, tracking targets is important and most approaches for multi-camera tracking are driven by the state-of-the-art visual object detection, tracking and re-identification methods. 
While single-camera tracking poses challenges like appearance, lighting, viewpoint and background variations and occlusions, multi-camera tracking with non-overlapping fields-of-view (FOV) poses a different challenge of re-identification of targets accross cmaeras. Since camera networks often have units that are spatially distant, transition times from one FOV to another may take several seconds or minutes or even longer depending on the scale of the camera network. Depending on network size and the cameras' FPS, these networks generate a deluge of video frames, which are potential query candidates for the re-identification module. For handling such volumes, scalable methods are of vital importance. One common approach is to select the potential camera feeds where the target is likely to be present. This approach can benefit both manual and automated surveillance as fewer frames need to be processed for tracking targets of interest. Along the same lines, we investigate \emph{camera selection decisions} to identify the most likely camera frame where the target may reappear at the next time instance.

The inter-camera target handovers are typically resolved using visual re-identification (Re-Id) techniques, where the current template of the target is matched against all target candidates in \emph{all candidate cameras}. Even for small camera networks with non-overlapping camera FOVs, this association problem becomes very challenging because of the non-deterministic and unknown time a target takes to transition between two non-overlapping FOVs. This uncertainty results in a large number of candidate frames, each with possibly many target candidates. Since most Re-Id or verification approaches work at an operating point chosen based on a fixed False Alarm Rate (FAR), the number of false alarms will depend on the number of frames processed for Re-Id. Re-Id false alarms could be very detrimental to the tracker's performance. Hence minimizing the number of frames that undergo a Re-Id query is critical to the tracking performance in camera networks, as well as reduce the computational complexity necessary to reduce the processing of frames not queried. An intelligent camera frame selection strategy could benefit both the accuracy and efficiency of a multi-camera target tracking system. 

In this paper, we highlight this important yet relatively unexplored problem of camera selection in multi-camera target tracking. Ideally, none of the camera frames should be selected for a Re-Id query during a target transition period. These target transition times are typically time-varying and are characterized by target speed, inter-camera distance and other processes, which nonetheless, can be modeled from the video data captured from the cameras. Thus, we propose to \emph{learn} a camera selection policy that intelligently schedules Re-Id queries to resolve inter-camera handovers. We design our approach in a manner that the learning strategy directly leverages the video data and does not depend upon the network topology. We will show experimentally that our proposed method makes a very few queries to the network as compared to the baseline and other competing methods used in the literature.


The likelihood of a target appearing at the next time step in one of the cameras is time-varying and depends on various non-deterministic factors like target speed, occlusion and others. Based on this observation, it is natural to model the camera selection problem for scheduling Re-Id queries as a Markov Decision Process (MDP), which was investigated in our initial work \cite{sharmaICAPS} by employing the Q-Learning method to exactly solve the MDP. However, exact methods are hard to scale for larger camera networks, which have larger state and action spaces. Therefore, in this paper, we present an extension of \cite{sharmaICAPS} and show that deep learning based approximate methods like Deep Q-Networks (DQN) \cite{deepQMnih} can be effectively used to scale up our camera selection approach to larger camera networks. In addition to the datasets used in \cite{sharmaICAPS} like NLPR-MCT \cite{nlpr} 
, we also evaluate the approximate approaches with larger camera networks like the Duke MTMC dataset \cite{mtmc}. The learned camera selection policy is used for inter-camera tracking (ICT) to generate an action that corresponds to waiting for the next time step by selecting a \emph{dummy camera} or selecting one of the real cameras to schedule a Re-Id query. Finally, the policy is learned directly from the videos captured from the camera units and does not assume the knowledge of the underlying network topology. Nonetheless, in our experiments, we observe that the policy implicitly learns the network topology anyway.

The specific contributions of this paper are: 

\begin{enumerate}
    \item We highlight the importance of camera selection decisions to enable accurate and efficient target tracking in a network of cameras. 
    
    \item We extend our approach reinforcement learning based intelligent querying in camera networks (using exact Q-learing~\cite{sharmaICAPS}) using deep learning. The deep learning based approximation enables camera selections for larger camera networks whereas the exact methods fail in larger state space. We also include a dense reward that helps to distinguish between states.
    
    \item We make modifications to our state vector to handle larger networks and use the time-limits \cite{time_limits} to handle indefinite transition times, while still maintaining the MDP formulation and learn the policy using DQN, an approximate method (Details in Sec. \ref{sec:mdpq}). 
    
    
    \item We demonstrate over multiple real-world datasets pertaining to both indoor and outdoor environments that the learned camera selection policy queries a very small number of frames with a small trade-off on the recall values. 
	
\end{enumerate}

The rest of the paper is organized as follows: Sec. \ref{sec:related} discusses prior work relevant to the problem of multi-camera tracking and Deep Reinforcement Learning. We present the details of our formulation and training procedure in Sec. \ref{sec:method} and show comparative results and empirical evaluation in Sec. \ref{sec:results}. Then we discuss the limitations in Sec. \ref{sec:discuss} and conclude in Sec. \ref{sec:conclusion}. 
\section{Related Works}
\label{sec:related}

\subsection{Visual Tracking in Camera Networks}
Prior works~\cite{overlapPlayer,overlapNetFlow,overlapShah} assumed overlapping cameras and find 3D coordinates of the target object for tracking. These works rely on camera calibration and network topology to derive the 3D coordinates. Other approaches for tracking using 3D coordinates are network flow problem~\cite{overlapNetFlow}, Kalman filter based tracking using the homography matrix~\cite{overlapShah}. The overlapping FOVs is too strong a constraint and have limited application in the real world. 

There have been efforts made to track a target in non-overlapping FOVs. Initially, the tracking task was performed for inter-camera tracking (ICT) to find camera handovers using affinity model of the target's appearance~\cite{affinity}, social grouping model~\cite{SGM}, using data association methods across multiple cameras~\cite{asso,assoRefSet,assoInv}. The other approaches formulate the tracking problem using graph based approaches~\cite{SGM,nlpr_graph}, contextual information~\cite{nlpr_usc}, spatio-temporal mapping between 3D coordinates~\cite{nooverlapSpaTemp}, and clique based formulation~\cite{clique,mtmc}. Many other works also incorporate target's travel time to estimate the transition time of camera handovers~\cite{traveltime}. These approaches perform target tracking in a unified way. However, a multiple camera network setup offer multiple challenges in terms of illumination variation, occlusions, and uncertain transition times that may have a time-varying distribution for different targets. In this direction, many related works incorporated appearance based template matching to track the target across the camera network. The appearance cues of the target were modeled in~\cite{santhoshAppear,overlapNetFlow,assoInv}. The appearance of the target is generally captured by color~\cite{overlapNetFlow} or texture~\cite{assoInv} features. To handle lighting variations across different cameras, color normalization~\cite{nlpr_usc}, brightness transfer functions~\cite{traveltime} were used. Spatio-temporal reasoning~\cite{overlapPlayer,affinity} and graph based methods~\cite{nlpr_graph} were applied over the appearance features to perform inter-camera tracking. In appearance based information, Bayesian inference~\cite{huang} was applied by integrating color and size of the object with the velocity and the arrival times of the target in two camera views. The approach was extended to more than two camera views using hidden variables~\cite{pasula} in the Bayesian formulation. Matei et al. ~\cite{matei}, on the other hand used a multi-hypothesis framework instead of a Bayesian model. Other approaches to multi-camera target tracking are using conditional random fields (CRF)~\cite{nlpr_crf}, global graph model using MAP association and flow graphs~\cite{nlpr_graph}. The state-of-the-art~\cite{nlpr_online} on NLPR dataset uses a two step framework to perform SCT (Single Camera Tracking) and ICT (Inter-Camera Tracking) separately. They incorporate multiple human appearance features along with segmentation using change point detection to perform SCT. They perform ICT by making a camera link model using a combination of appearance features and the distribution of transition time of the target between camera pairs. 

Apart from above related works, re-identification based approaches~\cite{REID, reid_survey} are prominent for template matching to associate different bounding box detection. All-pair template matching using re-identification and data association is found to be NP-hard~\cite{nphardTrack1,Chari2015OnPC} and hence multiple methods~\cite{deepcc,nlpr_crf} use time-consecutive frames to reduce the search complexity. ~\cite{deepcc} uses correlation clustering to trade off the computational cost. In contrast to these works, we propose a reinforcement learning based policy learning approach which selects a camera where the target is likely to be present at the next time step with the goal of reducing the search space for template matching (Re-Id). Our approach can be readily integrated with \emph{any} Re-Id approach as we only focus on the frame selection component.

Deep learning based approaches have shown superior template matching performance. For example, Ristani et al.~\cite{deepcc} proposed a weighted triplet loss for re-identification for better feature representation. They achieve multi-camera tracking using correlation clustering. However, their approach is restricted to tracking targets offline. In this paper, we will show that camera selection decisions are crucial to enable tracking in camera networks by comparing number of frames to be processed by various related methods. Moreover, our camera selection approach can benefit both online and offline target tracking in multi-camera networks.

\subsection{Deep Reinforcement Learning}
Many vision problems~\cite{MDPTrack,adnet,visionAppQlearn_ObjRec,visionApp2_AnytimeRecObj,deepQMnih} have been formulated using Markov Decision Process (MDP)~\cite{MDPBellman}. Formulating the tracking problem using MDP is effective because the agent learns to take actions sequentially, which implicitly model the target's motion. In our formulation, we have used one MDP for camera selection decision to enable single target tracking in multiple cameras which can easily be extended to tracking multiple targets by simultaneously running multiple policies. Deep-Q learning~\cite{deepQMnih} has shown human level performance in playing Atari games using visual frames. Such methods use one-step reward during the training process, however, n-steps reward~\cite{Sutton} can help in faster convergence by bootstrapping states for multi-step reward. Time limits~\cite{time_limits} in reinforcement learning has shown that randomizing the state vector after a time limit achieves better performance. Recently, deep reinforcement learning techniques were applied for visual object detection~\cite{RL_objDet} and tracking~\cite{adnet,tracking_onlineDecision,RL_end2end}. These approaches are applied for single object/target tracking in a single camera field-of-view. To our knowledge, we are the first to explore deep reinforcement learning for single target tracking in a camera network. We have shown that a policy learned using reinforcement learning can intelligently poll cameras to reduce the number of frames required for target's template matching. In our approach, we have used deep-Q learning~\cite{deepQMnih} to learn a policy to poll a camera frame at any time-step to look for the presence of the target.

\section{Proposed Methodology}
\label{sec:method}

In this section, we will provide the details of the system architecture and the reinforcement learning formulation for the camera selection decision problem. 

\begin{figure}
\centering
\includegraphics[width=1\textwidth]{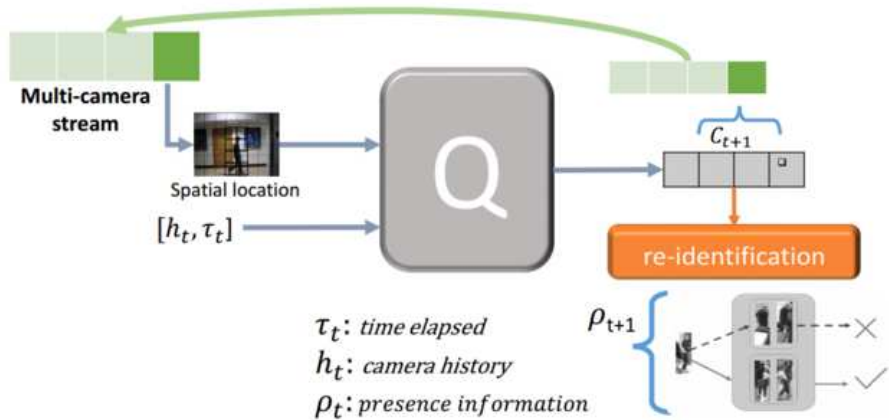}
\caption{The proposed architecture using reinforcement learning. The architecture shows two blocks, block \emph{Q} and block \emph{presence}. Block Q learns a policy to select a new camera using current state and block \emph{presence} verifies whether the target is present in the camera frame chosen. }
\label{fig:arch}
\end{figure}

\subsection{System Overview}
\label{sec:method_arch}
Figure ~\ref{fig:arch} shows our system architecture, which consists of two blocks: First, block Q which learns a policy $\pi$ to select the next camera where the target is likely to appear given target's current state. The second block verifies the presence of the target in the selected camera frame. In surveillance, this is usually done manually using human input or automatically using re-identification~\cite{reid_survey,REID} based approaches. We will name this second block, the \emph{presence} block. The \emph{presence} block takes as input the selected camera frame and will return $1$ if the target is present in the camera frame along with the corresponding bounding box location, otherwise it returns a $0$. As our focus is on learning the policy for camera selection, we being with the assumption that the presence block is perfect, and then investigate the impact of error in presence prediction. We achieve this by using ground truth labels for simulating a perfect Re-Id approach, and then induce random matching errors at different levels, in effect simulating outputs from Re-Id models at different levels of accuracy. This setting is followed in order to systematically evaluate the strength of our camera selection policy.

\begin{figure}[t!]
	\centering
	\includegraphics[width=12cm]{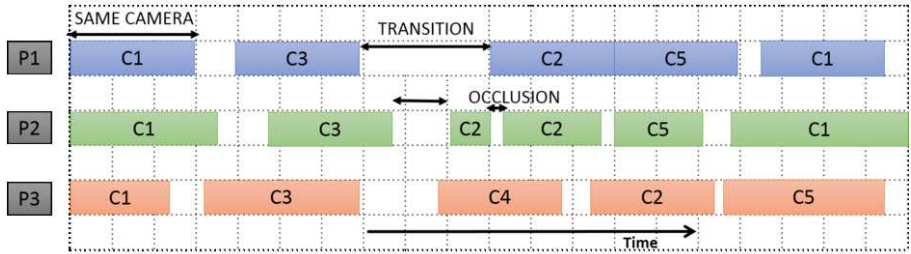}
	\caption{A example plot showing transitions by three target in a network of $5$ cameras. This example figure shows that a particular target's transition time is indefinite between camera handovers and there are occlusions in single camera trajectory. The figure shows that the transition time of different targets are not trivially related. }
	\label{fig:ex_transitions}
\end{figure}

The block Q, takes as input the current state (detailed in next subsection) and selects a camera index which will be polled to search the target using the presence block. The policy selects one of the $N+1$ actions, where $N$ is the number of cameras. The first $N$ actions correspond to each camera and the $N+1^{th}$ action is to be selected when the target is transitioning from one camera's FOV to another. The sequence of selected cameras gives the target's trajectory in terms of the cameras in which the target appears termporally. This is a non-trivial task due to the unknown and non-deterministic transition time of each target during camera transitions which also requires to correct any wrong selections made at previous timestamps. Consider the examples shown in Fig. ~\ref{fig:ex_transitions}, where the transition times between a pair of cameras are different for all targets. The policy is implemented using a neural network model depicted in Fig. ~\ref{fig:nnet}. The network parameters are learned using deep Q-learning~\cite{deepQMnih} with n-steps bootstrapping~\cite{Sutton}. In the subsequent subsection, we will provide details of the training and testing algorithm for camera selection decisions. 

\begin{algorithm}
\caption{Deep Q-network training procedure with n-step bootstrapping. $\pi$ is the policy to make camera selection decisions. \texttt{c,b} is the initial location of the target with $c$ as the current camera and $b$ as the corresponding bounding box location.}
\label{alg:train}
\begin{algorithmic}[1]
\Procedure{TRAIN}{$\texttt{c,b},\pi$}
\State \text{Initialize replay memory} $M$ \text{with capacity} $D$
\State \text{Initialize deep-Q network with random weights}

\State $h,\tau \gets \texttt{ZEROS}$\Comment{\small{Initialize history and time-elapse with ZEROS}}

\State $s\gets \texttt{initialState(c,b,h},\tau)$\Comment{\small{Concatenate location and history}}

\While {True}
    \State \texttt{With probability} $\epsilon$, \texttt{ choose action c uniformly at random, and with probability} $1-\epsilon$, \texttt{choose action using the policy in equation}~\ref{eq:pi}

    \State bb $\gets$ \texttt{getBoundingBox(c)}\Comment{\small{get the bounding box from presence block}}
    
    \If {\texttt{rand()} $< 0.5$}\Comment{\small{random steps are taken for exploration}}
        \State \texttt{rsteps} $\gets$ \texttt{randint(20)}
    \EndIf
    \If{bb is NOT EMPTY }\Comment{\small{update last seen observation vector if target is present in $c$}}
        \State $x_t$ $\gets$ \texttt{(c,bb)}
        \State $\tau\gets \texttt{ZERO}$
    \Else
        \State $\tau$ $+= rsteps$
    \EndIf
    \State $h\gets$ \texttt{updateHistory(h,c)}
    
    \State $s'\gets \texttt{f} (x_t,h,\tau)$\Comment{\small{observe the next state and reward}}
    \State $r \gets \texttt{getReward(s,c)}$
    
    \State \texttt{Append transition} $(s,c,s',r)$ \texttt{to replay memory M, pop last element if overflow}
    \If {$s'$ is terminal}
        break
    \EndIf
    \State $s\gets s'$
\EndWhile
    \State \texttt{Sample a random minibatch B from M}
    \State \texttt{For each sample} $(s_i,a_i,s_i',r_i)$ \texttt{in minibatch, compute the n-step target} $y_i = r_i+\gamma max_{a} Q(s_i',a)$ \Comment{\small{For brevity, target value is shown for 1-step reward}}
    \State \texttt{Update the Q-network using adam algorithm}~\cite{adam} \texttt{on the minibatch and repeat until convergence}
    
\State \textbf{return} $\pi$
\EndProcedure
\end{algorithmic}
\end{algorithm}

\subsection{Markov Decision Process and Q-learning}\label{sec:mdpq}

The goal of reinforcement learning is to learn a policy that decides sequential actions specific to the target's state by maximizing a cumulative reward function~\cite{Sutton}. Our system architecture uses deep Q-learning to learn a policy to make camera selection decisions. A decision problem can be formulated using a Markov Decision Process (MDP). 
The MDP is defined by the tuple $(S,A,f,R)$, where $S$ the set of states, $A$ is the set of actions, $f_a(s_t,s_{t+1})$ is the state transition function and $R(s,a)$ is the reward function that determines the reward that the environment provides by taking an action $a\in A$ at state $s\in S$. In an MDP, we learn a stochastic policy to decide the probability of an action given the current state of the agent. The environment then responds at next time with the next state (decided by the state transition function) and a reward (decided by the reward function). In real-world, both the state-transition function and the reward function can be stochastic. Given the MDP formulation, we can learn the policy using trial-and-error strategy. Our MDP formulation is given in the later part of this subsection. We define a state-action value function Q to estimate the expected value of the return for taking action $a$ in state $s$ by $Q(s,a)$. Return $R_t$ at time $t$ is defined as the discounted sum of future rewards:

\begin{equation}
    R_t = r_{t+1} + \gamma*r_{t+2} + \gamma^2*r_{t+3} + \dots = \sum_{k=0}^{\infty} \gamma^k r_{t+k+1}
\end{equation}

Where $\gamma$ is the discount factor which is typically included to make the return bounded. The estimates will then be used to make the action selection decision. We use state-action value function $Q^\pi(s,a)$ to learn the expected return starting at state $s$ and taking action $a$ and using policy $\pi$ for further time-steps (we will use $Q(s,a)$ in place of $Q^\pi(s,a)$ for all following text). The value function will tell us the expectation of how good (in terms of reward) the current state and action will result in future given the current policy. 

\begin{equation}
    Q(s,a) = \mathbb{E} \Big[\sum_{k=0}^{\infty} \gamma^k r_{t+k+1} | S_t=s, A_t=a \Big]
\end{equation}

The goal is to learn the optimal state-action function $Q^*$~\cite{Sutton}. The optimal Q-function $Q^*(s,a)$ can be learned using reinforcement learning techniques such as Q-learning, policy gradient, etc. We use Q-learning to learn an optimal Q-function because it is an off-policy and model-free algorithm i.e., an optimal policy can be learned by state-space exploration using trial-and-error and doesn't need an accurate state transition model. We have explained the learning procedure later in this subsection. 

We formulate the camera selections as a decision problem where each camera is considered as a separate action. 
As noted by related works~\cite{deepcc}, the target tracking over a camera network can be NP-hard for searching the target in all cameras and at all times and therefore, this becomes important to reduce the number of search operations while tracking a target across the multiple cameras. Selecting one camera for the search operation will reduce the need for searching across all cameras. The cameras in a typical camera network are deployed far apart and hence searching is pointless when the target is transitioning between cameras. To ensure this the policy learns to decide a null camera when the target is not visible in any of the camera. Therefore, the task is to learn a policy $\pi(s_t)$ at target's state $s_t$ which will give the probability of selecting a camera (equivalently selecting an action) given the current state i.e., $p(a_t|s_t)$, where $a_t$ is the action (or equivalently camera in the context of camera selection decisions). Such a policy can predict the period of visibility (when it is visible in any of the camera) of the target in the camera network, and the period of invisibility (when the target is not visible in any of the camera). We will show that this policy can be learned directly from data using the trial-and-error based approach i.e., by taking feedback from the environment. 

However, this problem doesn't map to the MDP directly because of the target's partial observability like occlusion from other targets or the target not present in the selected camera. For example, if the policy selects camera $c_i$ but the target is present in the FOV of camera $c_j$. The observation that the learning agent gets from the environment doesn't provide the target's state information and it makes the state non-Markov. Hence, we need to create a state vector from the noisy observations which is Markov (next state is independent of the previous state, given the current state). For partial observable environment, we can keep a history vector of the observations starting from the initial location to the current time which helps to estimate the next state. However, considering full history length in the state vector becomes intractable. Therefore, we need to create state from history which is Markov. 
In addition to the observations, we keep the action history and time elapsed in the state vector. To read more about the partial observable problem, readers are encouraged to read~\cite{Sutton}. The individual components of the state vector are defined in the following text: 

\textbf{State}: The state $s_t$ at time $t$ captures the observations of the target and the history of cameras $h_t$ selected by the policy, and time elapsed $\tau$. 
The individual elements of the state space are following:

\begin{enumerate}
    \item $x_t$: An observation of the target's location is its spatial location in a particular camera frame i.e., $(c,b)$ where $c$ is the camera index and $b$ is the bounding box in the camera $c$. If we keep last $3$ observations of the target then the next location can be estimated (for example, using kalman filter) and hence the last $3$ observations make the state vector Markov. 
    The last $3$ observations form the vector $x_t$. In which $c$ is encoded as a one-hot vector and $b$ is encoded by normalizing the bounding box location i.e., $x,y,w,h$. $(x,y)$ are the pixel coordinates of the upper left corner of the bounding box and $(w,h)$ are corresponding width and height respectively. The bounding box values are normalized by dividing the pixel coordinates by the corresponding image dimensions. 
    \item $h_t$: The action at next time-steps depends on the current action and the previous actions selected by the policy. Hence, we have included the previously selected actions to the state vector. $h_t$ represents the history of the cameras selected by the learned policy. The history of cameras is encoded as one-hot vector. 
    \item $\tau$: It captures the time elapsed since the target was last seen in any camera. This captures the time ticks since the target is not visible. Motivated from time-limits in reinforcement learning~\cite{time_limits}, we have included $\tau$ to work with indefinite transition times. For an infinite horizon problems, the time limits motivates that the state should be randomized after when the time-limit expires. Randomizing the state after time-limits achieves better performance. 

\end{enumerate}

\textbf{Actions}: The action $a_t$ at time $t$ is encoded by $N+1$ dimension vector, where $N$ is the number of cameras in the camera network. An optimal policy should select an action $a$ from the first $N$ actions when the target is visible in the camera index $a$. The action ${N+1}$ is selected when the policy selects no camera, i.e., the target is not visible in any of the camera. 

\textbf{State transition function}: After deciding an action $a_t$ at time $t$, the next state $s_{t+1}$ is decided by following state evolution function: 
\begin{equation}
    s_{t+1} = f(s_t,a_t)
\end{equation}

The function appends the one-hot encoding of the selected camera $c_t$ to the camera history vector. If the target is found in selected camera then last seen observation vector $x_t$ is updated by including new $(c,b)$ otherwise $\tau$ is incremented by $1$. 

\textbf{Reward}: The reward function $r(s,a)$ is defined for each state and action $a$ pair. In~\cite{sharmaICAPS}, we provided a binary reward function and here we use a dense reward. During training, this reward helps in knowing how long will it take to end the current handover. give At time $t$, it is following: 
\begin{equation}
 r(s_t,a) =
  \begin{cases}
    \frac{1}{T_c}       & \quad \text{if target will appear in selected camera c (=a) in time } T_c\\
    0.1  & \quad \text{if a=Cx is correct} \\
    -1  & \quad \text{otherwise}
  \end{cases}
\end{equation}

\textbf{Assumptions}: We assume that all the cameras of the camera network are uniquely identifiable and the camera network topology doesn't change during testing phase (the CCTV network infrastructure doesn't frequently change in the real world too).

\textbf{Policy}: The policy $\pi$ selects an optimal action from the learned Q-value functions. After learning, given the target state, it selects an optimal action using the learned Q-value function in-state $s_t$ as:
\begin{equation}
\label{eq:pi}
    \pi_t^*(s_t) = arg \max_{a} Q^*(s_t,a) 
\end{equation}

\textbf{Q-learning}: Q-learning is a temporal-difference (TD)~\cite{Sutton} learning algorithm which learns directly from state-space exploration without knowing a state-transition model. The Q-learning learns an optimal Q-value function by iteratively updating the values using the following bellman equation independent of the policy being followed: 
\begin{equation}
\label{eq:bellman_1step}
    Q(s_t,a_t) \Leftarrow  Q(s_t,a_t) +  \alpha \bigg( r(s_{t+1}) + \gamma \max_{a} Q(s_{t+1},a) - Q(s_t,a_t) \bigg)
\end{equation}

Where $\alpha$ is the learning rate, and $\gamma$ is the discount factor. At state $s_t$, the learning agent performs an action $a_t$ and then the environment responds with a new state $s_{t+1}$ and a reward value. An optimal Q-function should reflect the expected return for the state-action pair. Usually we start with a random policy and we explore the state-space by taking actions to update the value function about the goodness of the state-action pair. Sufficient exploration is essential for the Q-learning methods to update returns for a large number of state-action pair. In RL, we use epsilon-greedy exploration strategy~\cite{Sutton}. The update value considers the reward received for next one-step only but the one-step reward doesn't give the actual future reward during initial steps and the policy will not learn the right camera handover for larger transition times. Hence we are incorporating n-step rewards~\cite{Sutton} to update the value function. 

\textbf{Q-learning with n-step bootstrapping}:
The Q-learning update equation specified above updates the value function at next time using one step reward. In n-step reward, we update the value of a state after receiving rewards for n time steps. For example, taking $n=3$, would change the Q-learning bellman equation~\ref{eq:bellman_1step} to: 

\begin{equation}
\label{eq:bellman_nstep}
    Q(s_t,a_t) \Leftarrow Q(s_t,a_t) +  \alpha \bigg( r(s_{t+1}) +\gamma r(s_{t+2}) +\gamma^2 r(s_{t+3}) + \gamma^3 \max_{a} Q(s_{t+3},a) - Q(s_t,a_t) \bigg)
\end{equation}


\subsection{Camera Selection Decisions using Deep-Q Network}

Earlier in ~\cite{sharmaICAPS}, we proposed an exact RL method (the learned value function is stored in a table) for camera selection decisions where we discretized the state because of a very large state space but using deep learning we can learn features even from the continuous and larger state space. 
Neural networks were found to map the states to reward values in many related works~\cite{deepQMnih,alphaGo,lstmQ}. The parameters of the neural network can be updated using gradient descent based backpropagation algorithms~\cite{adam}. For all implementation of exact RL method, we have used a server machine with 128 GB RAM, 5GB GPU (Nvidia Tesla K20m) and Matlab-16B version. For implementation of neural networks, we have used a workstation with 8GB GPU (Nvidia GeForce GTX-1080), 16GB RAM and in pytorch. The exact RL method worked only for NLPR MCT datasets and goes Out-of-Memory (OM) for DukeMTMC dataset. 

\begin{figure}
\centering
\includegraphics[width=10cm]{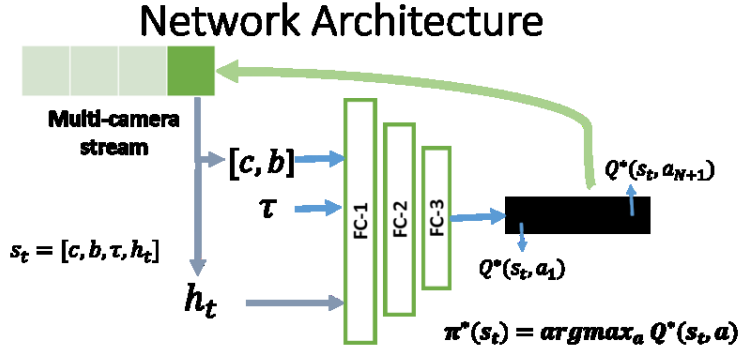}
\caption{The neural network model that learns the state-action values using Q-learning. The model learns a policy that makes the camera selection decisions. This is the implementation of the \emph{Q} block of the architecture shown in figure~\ref{fig:arch}.}
\label{fig:nnet}
\end{figure}

\textbf{Neural network model}: Our neural network model is shown in figure~\ref{fig:nnet}. For the neural network, we will find the optimal weights which will help the learning agent to get maximum reward. For the reward based learning, we have used deep Q learning~\cite{deepQMnih} algorithm to update the neural network weights based on the reward received from the environment. The first three hidden layers of the network have \textit{relu} activation and the last layer, outputs the Q-values corresponding to each individual action has linear activation. The output is a $N+1$ dimension vector, where $N$ is the number of cameras in the camera network. Each output corresponds to an action $a_i$ reflects the Q-value $Q(s,a_i)$ for the input state $s$. The action corresponding to maximum Q-value of the output layer is selected by the policy (equation~\ref{eq:pi}). The selected camera frame is then passed to the \emph{presence} block of the system to find the bounding box location of the target in the selected camera. The system then moves to the next state using the state-transition function and on the next state, the policy again selects an action using the Q-values predicted by the neural network. 

\textbf{Training procedure}: To train the network, we need to have the target labels corresponding to each input. Deep-Q learning~\cite{deepQMnih} algorithm uses the return at each step as the target label. The output of the network at state $s_t$ is $Q(s_t,a_t)$ $\forall a_t\in A$ and the corresponding target is the discounted future reward for $n$-steps. For simplicity, taking $n=1$, the target $y_t$ for state $s_t$ after receiving a reward $r_{t+1}$ from environment is $r_{t+1} + \gamma max_{a} Q(s_{t+1},a)$. We have used mean-square error to compute loss at each time-step. Hence, when action $a_i$ is taken at state $s_t$, the loss (corresponding to action $a_i$) can be written as:

\begin{equation}
    L(s_t,a_i) = \big(Q(s_t,a_i) - (r_{t+1} + \gamma max_{a} Q(s_{t+1},a))\big)^2
\end{equation}

The loss term for actions other than $a_i$ will be zero (there are $N+1$ actions). In RL, the term in brackets is also known as TD (Temporal Difference) error. In the loss for n-step bootstrapping, we replace the next (one) step reward with the n-step return. The step by step training procedure is shown in algorithm~\ref{alg:train}.

\begin{figure*}
     \centering
     \subfloat{\includegraphics[width=5.5cm]{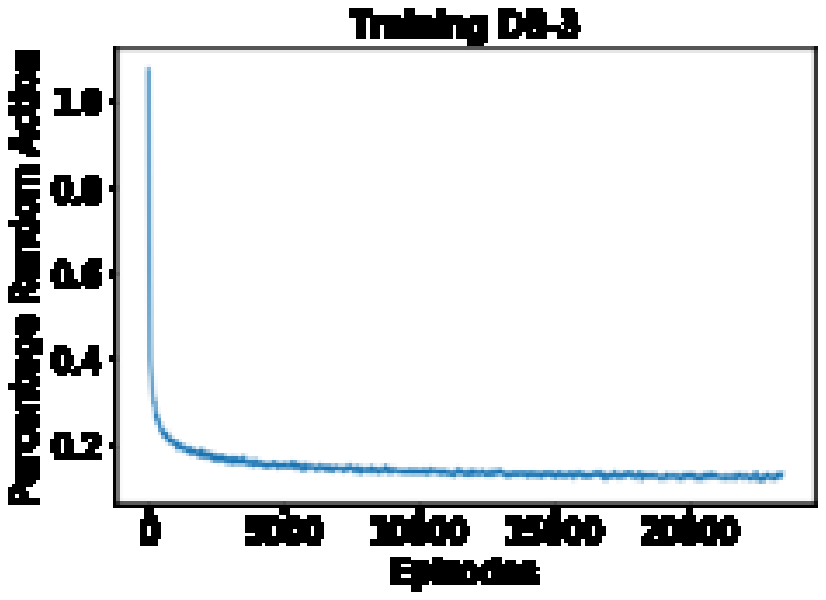}   }
     \subfloat{\includegraphics[width=5.8cm]{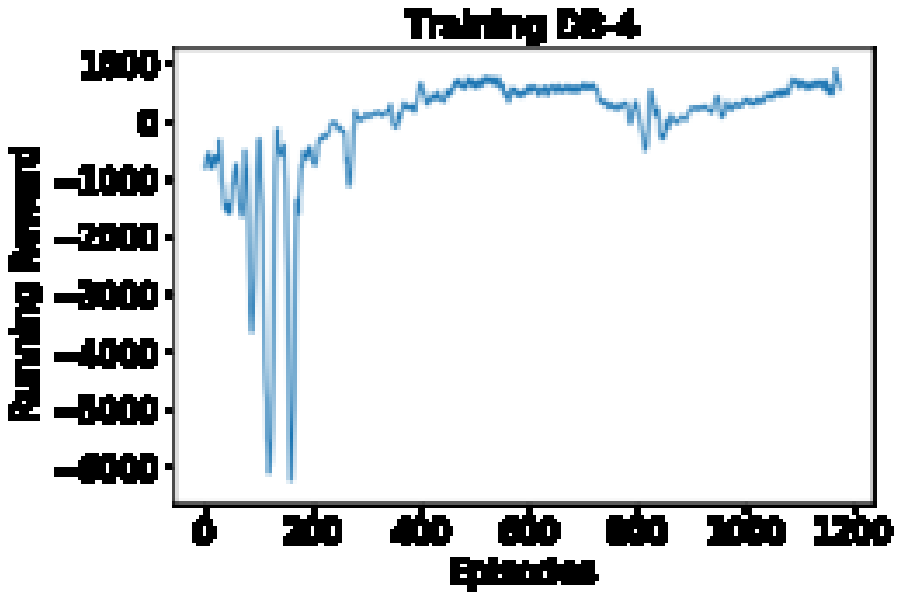}}
     \caption{Analysis of training strategy. First, shows the varying epsilon value during training on NLPR DB-3. Second, the running reward during the training on NLPR DB-4. }
     \label{fig:analysis_randAction}
\end{figure*}

Note that the training procedure is same irrespective of whether the target is inside a camera field-of-view or transitioning between cameras. For training the neural network, we initialized the state vector with the initial location of the target and history vector to all zeros. 
The selected action (camera index) is then used to verify the presence of the target (see section~\ref{sec:method_arch}). The state is accordingly updated using the state transition function. At any particular time, a target can see occlusion during SCT (Single Camera Tracking) and hence to simulate such cases, we have included short random jumps and hence $\tau$ increments by the value of the random jump or by $1$ when presence block cannot find the target. If the target is found, $\tau$ is set to $0$. Each transition is stored in a replay memory until the end of the episode. When episode ends, a small minibatch is sampled randomly from the replay memory for backpropagation using adam~\cite{adam}. The training process is repeated until convergence (when the reward received in each episode saturates). Instead of fixing a value for the epsilon in epsilon greedy exploration, we start with a value of $1$ and decrements it as training progresses. The epsilon is set using $1/log(epoch\_number)$. At later training epochs, the policy's decision was used as the epsilon values reaches a minimum as shown in the figure~\ref{fig:analysis_randAction}. The second plot of the figure shows that the reward saturates at later training epochs. 


\section{Evaluation and Results}
\label{sec:results}
In this section, we present details of the datasets used, the evaluation metric and the experimental results of the proposed architecture on the used datasets. 

\begin{table}
\caption{Details of the datasets used for performance evaluation. The table shows the number of cameras ({\#}Cameras), duration of the videos, frame rate (FPS) and the number of people ({\#}People) captured in each dataset.}
\label{tab:data_descr}
\centering
 \begin{tabular}{|c c c c c c|} 
 \hline
  & NLPR\_Set1 & NLPR\_Set2 & NLPR\_Set3 & NLPR\_Set4 & DukeMTMC \\ [0.5ex] 
 \hline\hline
 {\#}Cameras & 3 & 3 & 4 & 5 & 8 \\ 
 Duration & 20 min & 20 min & 3.5 min & 24 min & 1hr 25min \\ 
 FPS & 20 & 20  & 25 & 25 & 60 \\ 
 {\#}People & 235 & 255  & 14 & 49 & 2834  \\ 
 \hline
 \end{tabular}
\end{table}

\subsection{Dataset and Evaluation Metric}
\label{sec:data}
\begin{figure}
	\centering
	\includegraphics[height=4.9cm]{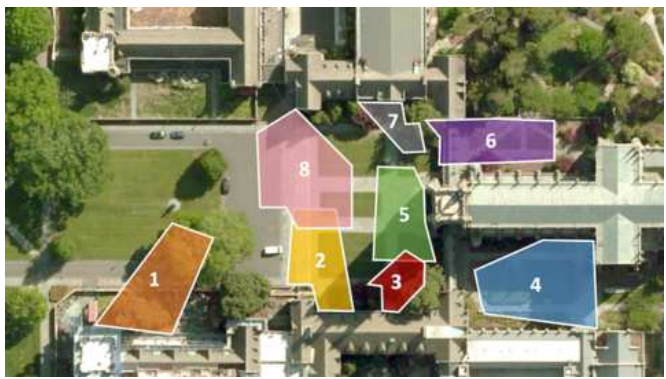}
	\caption{Top view of DukeMTMC dataset~\cite{deepcc}. The figure shows the top view of the camera network with field-of-view of all the eight cameras. The camera network is deployed in Duke university campus. }
	\label{fig:duke_topo}
\end{figure}

\emph{Dataset}: We have used NLPR\_MCT data set~\cite{nlpr} and DukeMTMC~\cite{mtmc} dataset to evaluate the proposed architecture for camera selections in multi-camera network for single target tracking. The NLPR\_MCT dataset consists of four sub-datasets each having $3-5$ cameras with a resolution of $320\times 240$. Details of the dataset are given in Table~\ref{tab:data_descr}. The dataset comprises cameras installed in both indoor and outdoor environments with significant illumination variation across different cameras. The set-1 and set-2 have the same environment and network topology. The set-3 was captured in an office building, and the set-4 was captured in a parking area. We learn a separate policy for set-3, set-4, and set-1. Since the camera network in set-2 is same as set-1, we use the same policy for both subsets. The DukeMTMC dataset consists of $8$ cameras deployed in Duke University campus. To date, DukeMTMC dataset is the benchmark dataset for multi-target multi-camera (MTMC) tracking. The details of the dataset are given in table~\ref{tab:data_descr}. It is difficult to identify the correct topology of the camera network with both overlapping and non-overlapping FOVs, for example in the case of the DukeMTMC dataset. The top view of the camera topology of DukeMTMC dataset is shown in figure~\ref{fig:duke_topo}.

The training and the testing sets are constructed from each datasets by randomly selecting half the people for the training and the remaining half for testing. 
However, the evaluation benchmark of DukeMTMC dataset doesn't provide platform for camera selection performance and hence to train the policy and to evaluate the performance, we have divided the available training set into two parts by splitting person identities in two sets. Therefore, for camera selection decisions, we are reporting performance on the sub-part of the actual training set. The two sets contain mutually exclusive person identities. We expect the policy to implicitly learn the network topology, and so long as the network is static, the policy should work for all new, unseen target individuals. Typically, CCTV network topologies in the real-world are seldom modified. 

We define evaluation metrics over the entire sequence of frames generated by the camera network. The sequence is indexed by time-steps corresponding to the time of frame capture for the cameras. Since the cameras operate on the same frame rate for a given subset, we can ignore any synchronization errors without any significant impact on the camera selection and tracking performance. 

\emph{Evaluation Metric}: 
To evaluate the camera selection performance, we report camera selection accuracy, precision and recall computed over the entire sequence of each subset. 
In order to consider instances when the target is not visible in any of the cameras, we introduce a dummy \emph{null} camera and denote it by $ C_\times $. Given a target, let the ground truth sequence of cameras in which it appears be contained in the vector $g$ and sequence of cameras polled by the policy be in vector $p$ with the $ i^{th} $ element indicated using a subscript. The Accuracy (A), precision (P) and recall (R) are defined as following for a single target
\begin{align}
\label{eqn:apr}
    A & = \frac{\sum_i(p_i==g_i)}{\texttt{Length}(g)}\\
    P & = \frac{\sum_i((p_i==g_i)~\land~ (p_i!=C_\times))}{\sum_i(p_i!=C_\times)}\\
    R & = \frac{\sum_i((p_i==g_i~\land~g_i!=C_\times)}{\sum_i(g_i!=C_\times)}
\end{align}
The final value for each of these metrics is reported as an average computed over all targets. Along with $A, P, R$, we also report number of frames polled ($ F $) during an inter-camera transition of the target. It is defined as
\begin{align}
\label{eq:f}
F = & \sum_i\left((g_i==C_\times)~\land~(p_i!=C_\times)\right) \quad + \\  
\nonumber & \sum_i\left((p_i!=g_i)~\land~(g_i!=C_\times)~\land~(p_i!=C_\times)\right) 
\end{align}
$F$ is an important measure because with a large number of frames polled, the chance of false alarms during a re-identification query as well as the computational complexity is substantially increased. We perform evaluation in two parts, one for ICT alone and another for ICT along with SCT. For ICT alone case, we do not consider the frames when the target was seen in a single camera field-of-view. 
We also evaluate the overall performance of target tracking in a camera network, when our camera selection policy is used for ICT. We use the standard Multi-Camera Tracking Accuracy (MCTA), which gives a single scalar value for all components involved in multi-camera tracking, i.e., F1-score for detection, number of target handovers for single camera tracking, and the number of handovers in inter-camera tracking. The metric is defined as
\begin{equation}
MCTA = \underbrace{\left(\frac{2P_TR_T}{P_T+R_T}\right)}_{F1-score} \underbrace{\left(1-\frac{\sum_t \mu_{t}^{s}}{\sum_t tp_{t}^{s}}\right)}_{within-camera} \underbrace{\left(1-\frac{\sum_t \mu_{t}^{c}}{\sum_t tp_{t}^{c}}\right)}_{cross-camera}
\end{equation}
where $P_T$ is the precision, $R_T$ is recall for target IDs. The number of target-ID mismatches at time $t$ is given by $\mu_t$ and $tp_t$ is the number of true positives in a single camera at time $t$. The superscripts $ s $ and $ c $ denote the single camera tracking (SCT) or cross-camera tracking (ICT) scenario. Readers are requested to see \cite{mtmc,nlpr} for details about the MCTA metric.


We have proposed a single target tracking approach that tracks the given target across multiple cameras whereas the related approaches on the benchmark datasets are multi-target multi-camera. To make a fair comparison with related approaches, we have created a multi-target version of our algorithm. To compute multi-target tracking results, we are running multiple pipelines of our approach for multiple targets. In our approach, the tracking performance of one target does not depend on another and hence the approach can be easily extended to multi-target tracking problem. 



    
    
    

\begin{algorithm}
\caption{Camera selection decisions using deep Q learning.}
\label{alg:test}
\begin{algorithmic}[1]
\Procedure{SelectionDecisions}{$\texttt{c,b},\pi$}

\State $h,\tau\gets \texttt{ZEROS}$\Comment{\small{Initialize history and time-elapse with ZEROS}}
\State $s\gets \texttt{initialState}(c,b,h,\tau)$\Comment{\small{Create initial state using history and location}}
\While {the video sequence ends}
    \State $c = \texttt{argmax} (\pi(s))$\Comment{\small{Choose an action using the learned policy}}
    \State \texttt{Select a random c, if} $\tau$ \texttt{ reaches the max time-limit }
    
    \State $\texttt{b}\gets \texttt{get the bounding box location using presence block}$
    \If { \texttt{b} is not empty }
        \State \texttt{Update} $x_t\gets  \texttt{(c,b)}$ \texttt{and} $\tau\gets \texttt{ZERO}$ 
    \Else \State $\tau += 1$ 
    \EndIf
    
    \State $h\gets \texttt{updateHistory(h,c)}$\Comment{\small{Update history at every time-step}}
    \State $s\gets f(x_t,h,\tau)$\Comment{\small{Observe next states}}
    \State \texttt{Append (c,b) to trajectory}
\EndWhile

\State \textbf{return} $\texttt{trajectory}$
\EndProcedure
\end{algorithmic}
\end{algorithm}





\subsection{Camera Selection Performance of the Learned Policy}


\begin{table}[t]
\caption{Table is showing camera selection accuracy (A), precision (P) and recall (R) for the proposed method and baseline approaches for NLPR dataset for the case of Inter-Camera Tracking (ICT). }
\label{tab:csa_nlpr_ict}
\centering
\begin{tabular}{|c|c|c|c|c|c|c|}
    \hline
    & $A$ & $P$ & $R$ & $A$ & $P$ & $R$ \\ \hline
    &\multicolumn{3}{|c|}{NLPR DB-1}&\multicolumn{3}{|c|}{NLPR DB-2} \\ \hline
    
    Exhaustive &0.025 &0.008 &1.0 &0.019 &0.007 &1.0 \\ \hline
    Neighbor &0.025 &0.013 &1.0 &0.019 &0.009 &1.0 
    \\ \hline
    Gaussian &0.435 &0.215 &0.127 &0.40 &0.16 &0.195  \\ \hline
    Exact RL &0.85 &0.042 &0.31 &0.86 &0.037 &0.31 \\ \hline
    Deep RL & 0.44 & 0.026 & 0.73 & 0.45 & 0.02 & 0.73 \\
    \hline
    
    & \multicolumn{3}{|c|}{NLPR DB-3}& \multicolumn{3}{|c|}{NLPR DB-4} \\ \hline 
    Exhaustive & 0.008 & 0.002 &1.0 & 0.017 & 0.003 &1.0  \\ \hline
    Neighbor & 0.008 & 0.003 &1.0 & 0.017 & 0.006 &1.0\\ \hline
    Gaussian & 0.36 & 0.007 &0.571 & 0.33 & 0.0078 &0.168 \\ \hline
    Exact RL & 0.685& 0.026& 0.929 & 0.519& 0.027&0.808   \\ \hline
    Deep RL & 0.58 & 0.02 & 0.88 & 0.76 & 0.03 & 0.83  \\ \hline
    
\end{tabular}
\end{table}

\begin{table}[t]
\caption{Table is showing camera selection accuracy (A), precision (P) and recall (R) for the proposed method and baseline approaches for NLPR dataset for the case of both ICT and SCT together. }
\label{tab:csa_nlpr_both}
\centering
\begin{tabular}{|c|c|c|c|c|c|c|}
    \hline
    & $A$ & $P$ & $R$ & $A$ & $P$ & $R$ \\ \hline
    &\multicolumn{3}{|c|}{NLPR DB-1}&\multicolumn{3}{|c|}{NLPR DB-2} \\ \hline
    Exhaustive &0.72 &0.24 &1.0 &0.65 &0.22 &1.0 \\ \hline
    Neighbor &0.72 &0.36 &1.0 &0.65 &0.32 &1.0 \\ \hline
    Exact RL &0.91 &0.95 &0.83 &0.88 &0.94 &0.78 \\ \hline
    Deep RL & 0.84 & 0.76 & 0.90 & 0.80 & 0.69 & 0.84 \\ \hline
    
    &\multicolumn{3}{|c|}{NLPR DB-3}&\multicolumn{3}{|c|}{NLPR DB-4} \\ \hline 
    
    Exhaustive & 0.42 & 0.10 &1.0 & 0.56 & 0.11 &1.0 \\ \hline
    Neighbor & 0.42 & 0.14 &1.0 & 0.56 & 0.18 &1.0 \\ \hline
    Exact RL & 0.76& 0.64& 0.86 & 0.77& 0.61&0.91 \\ \hline
    Deep RL & 0.73 & 0.60 & 0.88 & 0.93 & 0.73 & 0.84 \\ \hline
\end{tabular}
\end{table} 

\begin{table}[t]
\caption{Table is showing camera selection accuracy (A), precision (P) and recall (R) for the proposed method and baseline approaches for DukeMTMC dataset for both ICT alone and ICT-SCT together. The Gaussian approach is not defined for SCT+ICT case. In the table, OM signifies Out-of-Memory error. }
\label{tab:csa_duke}
\centering
\begin{tabular}{|c|c|c|c|c|c|c|}
    \hline
    &\multicolumn{3}{|c|}{ICT alone}&\multicolumn{3}{|c|}{SCT + ICT} \\ \hline
    
    & $A$ & $P$ & $R$  & $A$ & $P$ & $R$ \\ \hline
    Exhaustive & $9.6*10^{-4}$ & $1.2*10^{-4}$ & 1.0 & 0.334 & 0.042 &1.0 \\ \hline
    Neighbor & $9.6*10^{-4}$ & $2.4*10^{-4}$ &1.0  & 0.334  & 0.042  &1.0 \\ \hline
    Gaussian & 0.26 & $1.9*10^{-4}$ & 0.58 & - &- &- \\ \hline
    Exact RL & OM & OM & OM & OM & OM & OM \\ \hline
    Deep RL & 0.81 & $6.7*10^{-3}$ & 0.74 & 0.869 & 0.49 & 0.768  \\ \hline
    
\end{tabular}
\end{table} 

In this subsection, we will describe the performance of the learned policy for camera selection decisions. There are two cases for tracking a target in a camera network. First, ICT (Inter-Camera Tracking) where the task is to identify the correct camera handovers that the target performs. Second, is SCT+ICT (Single Camera Tracking + ICT) where the task is to identify the correct cameras when the target is moving in a single camera field-of-view along with the camera handovers. 

To perform the experiment, we have initialized the initial state of the target with its initial location with history vector being all zeros. At each time-step, the learned policy selects a camera index where the target is likely to be present. The selected camera is then queried to identify whether the target is present in the selected camera field-of-view. The presence of the target is used to locate its spatial location (bounding box) in the selected camera frame. For surveillance, this task is usually performed by human agents who continuously watch the camera feed. Alternatively, this task can be achieved by re-identification based methods to automatically identify the presence of the target. Such methods use visual template matching to re-identify an object in different camera feeds given the visual template of the target. To evaluate the camera selection decisions, we use correct presence of the target from the ground truth data. We make this simplifying choice in this experiment to eliminate the uncertainty introduced due to the re-identification performance. The policy continues polling of cameras until the target exits the camera network or the sequence terminates. The complete procedure to perform target tracking using the learned policy is shown in the algorithm~\ref{alg:test}. For infinite horizon problems, time limits~\cite{time_limits} in reinforcement learning have shown on various applications that randomizing the state vector (even during testing) after a time period provides better performance because larger time steps may end up in a bad state. Randomizing the state vector will help the policy to select actions from another state and eventually results in better performance. Similarly, in our case, when $\tau$ reaches a predefined maximum value, we select a random camera index to update the state vector and let the policy continue from that point to make camera selection decisions. For example, for NLPR DB-3, without using time limits, we got camera selection accuracy of $0.69$ whereas by setting a time limit of $250$ time-steps we got an accuracy of $0.73$. We observed similar case of other datasets and used a different time limit for all datasets. All further results are reported with time limits of $800$ for NLPR DB-1 and 2, $250$ for NLPR DB-3, $500$ for NLPR DB-4, and $600$ for DukeMTMC dataset. 

Metrics like accuracy, precision and recall encapsulate overall performances and allow comparative analysis as shown in Table~\ref{tab:csa_nlpr_ict},~\ref{tab:csa_nlpr_both}~and~\ref{tab:csa_duke} which reports the camera selection performance on each dataset. Table~\ref{tab:csa_nlpr_ict} shows accuracy (A), precision (P), and recall (R) 
for NLPR MCT dataset for ICT case only. Table~\ref{tab:csa_nlpr_both} shows A, P, R for NLPR MCT dataset for both SCT and ICT and Table~\ref{tab:csa_duke} shows the camera selection decision performance for DukeMTMC dataset for both cases, ICT alone and SCT and ICT together. 

In addition to the proposed policy's performance, 
we are comparing the camera selection performance of the policy with three baseline approaches used in related works. The \emph{Exhaustive} approach is a brute-force approach which polls each camera at all time steps until the target is found in one of the cameras. The table shows that it has $100\%$ accuracy but poor precision. The \emph{Neighbor} approach assumes that the camera network topology is known and searches the target by polling only in the neighboring cameras. Approaches proposed in~\cite{SGM,nlpr_crf} searches the target in the adjacent cameras and hence process the same number of frames as the \emph{neighbor search} approach. Along with these two approaches, we also compare camera selection performance with a method proposed in~\cite{nlpr_online}. The approach proposed in~\cite{nlpr_online} first estimates the distribution of the camera transitions assuming the fact that the multiple targets generally follow same paths and then samples a transition time to reduce the number of frames to be processed. They estimate a Gaussian distribution and hence we named this approach as \emph{Gaussian}. After the transition time, they start searching the target in cameras using a camera link model which will link different cameras having a path for transition. We repeated their experiment by estimating a Gaussian distribution from the train set and sampling a transition time for each person in the test set. The camera link model is used as set of neighboring cameras. The metrics computed in each table are reported for two cases: For ICT, the metrics are computed using equation (\ref{eqn:apr}), but only using the time instances when the target is transitioning from one camera to the other. In case of SCT + ICT, the entire sequences are used. As expected, we see that the proposed policy has better precision than the other competing approaches. The Gaussian method is excluded in case of SCT + ICT, as the distribution is only defined for the ICT case. While the A, P and R measures indicate the overall performance of camera selection, a confusion matrix shows the pairwise miss-classification in camera selection. Based on the cameras being polled by our policy at various time steps, we report a confusion matrix for DukeMTMC dataset as shown in Table~\ref{tab:confmat}. Our previous implementation in~\cite{sharmaICAPS} using Q-learning goes out of memory for this dataset due to a very large state space. The confusion matrix is computed using deep learning based approximation of the Q-learning algorithm. 

\begin{table}[h!]
\caption{Table is showing confusion matrix of the camera selections made by the proposed policy for DukeMTMC dataset. Rows are the ground truth cameras ($GT$) and columns are the cameras polled by the policy. Values are percentages rounded off to third decimal.}
\label{tab:confmat}
\centering
\begin{tabular}{|c | c c c c c c c c c|}
\hline
$\downarrow GT$ & $C_1$ & $C_2$ & $C_3$ & $C_4$ & $C_5$ & $C_6$ & $C_7$ & $C_8$ & $C_\times$\\ [0.5ex] 
\hline\hline

$C_1$ & 0.618& 0.137& 0.005& 0.005& 0.011& 0.005& 0.005& 0.005& 0.210 \\
$C_2$ & 0.012& 0.689& 0.006& 0.006& 0.015& 0.006& 0.006& 0.005& 0.254 \\
$C_3$ & 0.005& 0.004& 0.540& 0.011& 0.005& 0.009& 0.004& 0.004& 0.419 \\
$C_4$ & 0.000& 0.000& 0.000& 0.994& 0.000& 0.000& 0.000& 0.000& 0.005 \\
$C_5$ & 0.005& 0.004& 0.004& 0.004& 0.521& 0.004& 0.004& 0.005& 0.448 \\
$C_6$ & 0.003& 0.003& 0.003& 0.003& 0.003& 0.877& 0.004& 0.003& 0.102 \\ 
$C_7$ & 0.004& 0.003& 0.004& 0.004& 0.004& 0.004& 0.359& 0.004& 0.614 \\ 
$C_8$ & 0.000& 0.000& 0.000& 0.000& 0.000& 0.000& 0.000& 0.928& 0.071 \\
$C_\times$ & 0.022& 0.022& 0.022& 0.022& 0.022& 0.023& 0.022& 0.022& 0.825\\

\hline
\end{tabular}
\end{table}

\begin{figure*}
\centering
\includegraphics[width=12.5cm]{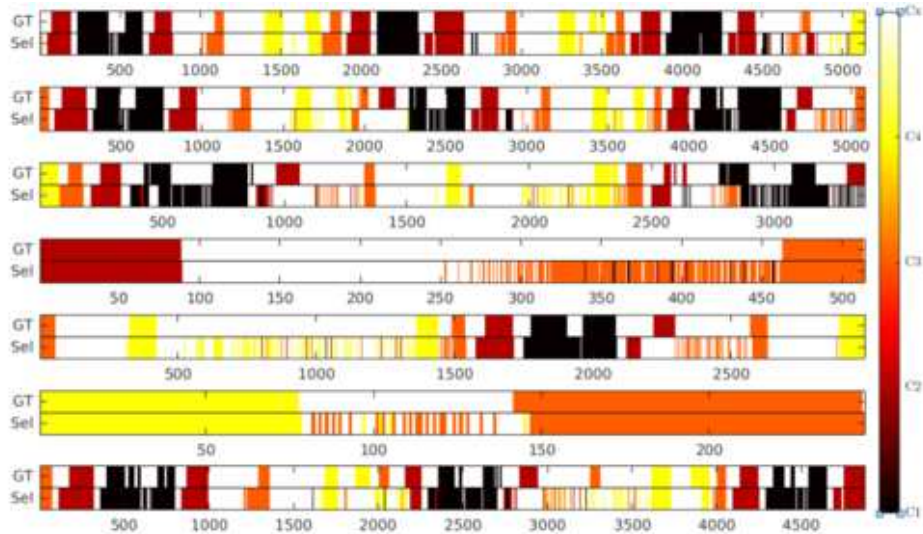}
\caption{The figure shows the transitions for $7$ targets in the testing set of dataset-3. On y-axis, $GT$ is the sequence of cameras in ground-truth, $Sel$ is the sequence of cameras polled by the policy. Horizontal axis is the time. White color is the length of the transition during camera handovers and colorbar depicts the camera numbers in the plot.  }
\label{fig:tran_traj_db3}
\end{figure*}

\begin{table}[h!]
\caption{The table is showing average MCTA values for inter-camera tracking (ICT) on the test set of NLPR\_MCT dataset. The related approaches are multi-camera multi-target tracking approaches taken from the benchmark dataset~\cite{nlpr}. The last $10$ rows show the MCTA values for the proposed approach with simulated re-identification errors from $0\%$ to $20\%$ for both Exact RL and Deep RL implementations. }
\label{tab:MCTA_ict}
\centering
 \begin{tabular}{|c| c c c c |}
 \hline
 & \multicolumn{4}{|c|}{\textbf{Inter-camera tracking (ICT)}} 
 \\ [0.5ex] \hline
  Approach & DB-1 & DB-2 & DB-3 & DB-4 
  \\ [0.5ex] 
 \hline\hline
\cite{nlpr_usc} & 0.9152 & 0.9132 & 0.5163 & 0.7152 \\ 
\cite{nlpr} & 0.7425 & 0.6544 & 0.7369 & 0.3945\\ 
\cite{nlpr_cripac} & 0.6617 & 0.5907 & 0.7105 & 0.5703 \\ 
\cite{nlpr_online} & 0.9610 & 0.9264 & 0.7889 & 0.7578 \\ 
\cite{nlpr_graph} & 0.835 & 0.703 & 0.742 & 0.385\\ 
 Exact RL-0 & 0.8210 & 0.7498 &0.9099 & 0.8993\\ 
 Deep RL-0 & 0.9016 & 0.8741 & 0.9038 & 0.8074 \\ 
 Exact RL-5 & 0.8188 & 0.7481 &0.8766 & 0.8137 \\ 
 Deep RL-5 & 0.7869 & 0.6994 & 0.6971 & 0.6118 \\ 
 Exact RL-10 & 0.8219 & 0.7511 &0.8848 & 0.7140 \\ 
 Deep RL-10 & 0.7293 & 0.5985 & 0.4390 & 0.5673 \\ 
 Exact RL-15 & 0.8171 & 0.7468 &0.7862 & 0.7128\\ 
 Deep RL-15 & 0.7043 & 0.5147 & 0.3658 & 0.3946 \\ 
 Exact RL-20 & 0.8203 & 0.7519 &0.7101 & 0.6625 \\ 
 Deep RL-20 & 0.6543 & 0.4540 & 0.3516 & 0.4680 \\ 
 \hline
 \end{tabular}
\end{table}

\begin{table}[h!]
\caption{The table is showing average MCTA values for SCT and ICT together case on the test set of NLPR\_MCT dataset. The related approaches are multi-camera multi-target tracking approaches taken from the benchmark dataset~\cite{nlpr}. The last $10$ rows show the MCTA values for the proposed approach with simulated re-identification errors from $0\%$ to $20\%$ for both Exact RL and deep RL implementation. }
\label{tab:MCTA_sct_ict}
\centering
 \begin{tabular}{|c| c c c c |}
 \hline
 & \multicolumn{4}{|c|}{\textbf{SCT + ICT}} 
 \\ [0.5ex] \hline
  Approach & DB-1 & DB-2 & DB-3 & DB-4  
  \\ [0.5ex] 
 \hline\hline
\cite{nlpr_usc} & 0.8831 & 0.8397 & 0.2427 & 0.4357 \\ 
\cite{nlpr} & 0.7477 & 0.6561 & 0.2028 & 0.2650\\ 
\cite{nlpr_cripac} & 0.6903 & 0.6238 & 0.0848 & 0.1830\\ 
\cite{nlpr_graph} & 0.8525 & 0.7370 & 0.4724 & 0.3778 \\ 
 Exact RL-0 & 0.8235 & 0.7503 & 0.9134 & 0.9118 \\ 
 Deep RL-0 & 0.9018 & 0.8806 & 0.9058 & 0.7871 \\ 
 Exact RL-5 & 0.7778 & 0.7064 & 0.7949 & 0.7338 \\ 
 Deep RL-5 & 0.6654  & 0.5585 & 0.2210 & 0.4624 \\ 
 Exact RL-10 & 0.7355 & 0.6635 & 0.6791 & 0.6769\\ 
 Deep RL-10 & 0.5846 & 0.4184 & 0.1333 & 0.3660 \\ 
 Exact RL-15 & 0.7004 & 0.6160 & 0.6229 & 0.5879\\ 
 Deep RL-15 & 0.5123 & 0.3130 & 0.1176 & 0.3084 \\ 
 Exact RL-20 & 0.6281 & 0.5323 & 0.5541 & 0.5288\\ 
 Deep RL-20 & 0.4096 & 0.2194 & 0.1196 & 0.2324 \\ 
 \hline
 \end{tabular}
\end{table}

\begin{figure*}
\centering
\includegraphics[width=10.5cm]{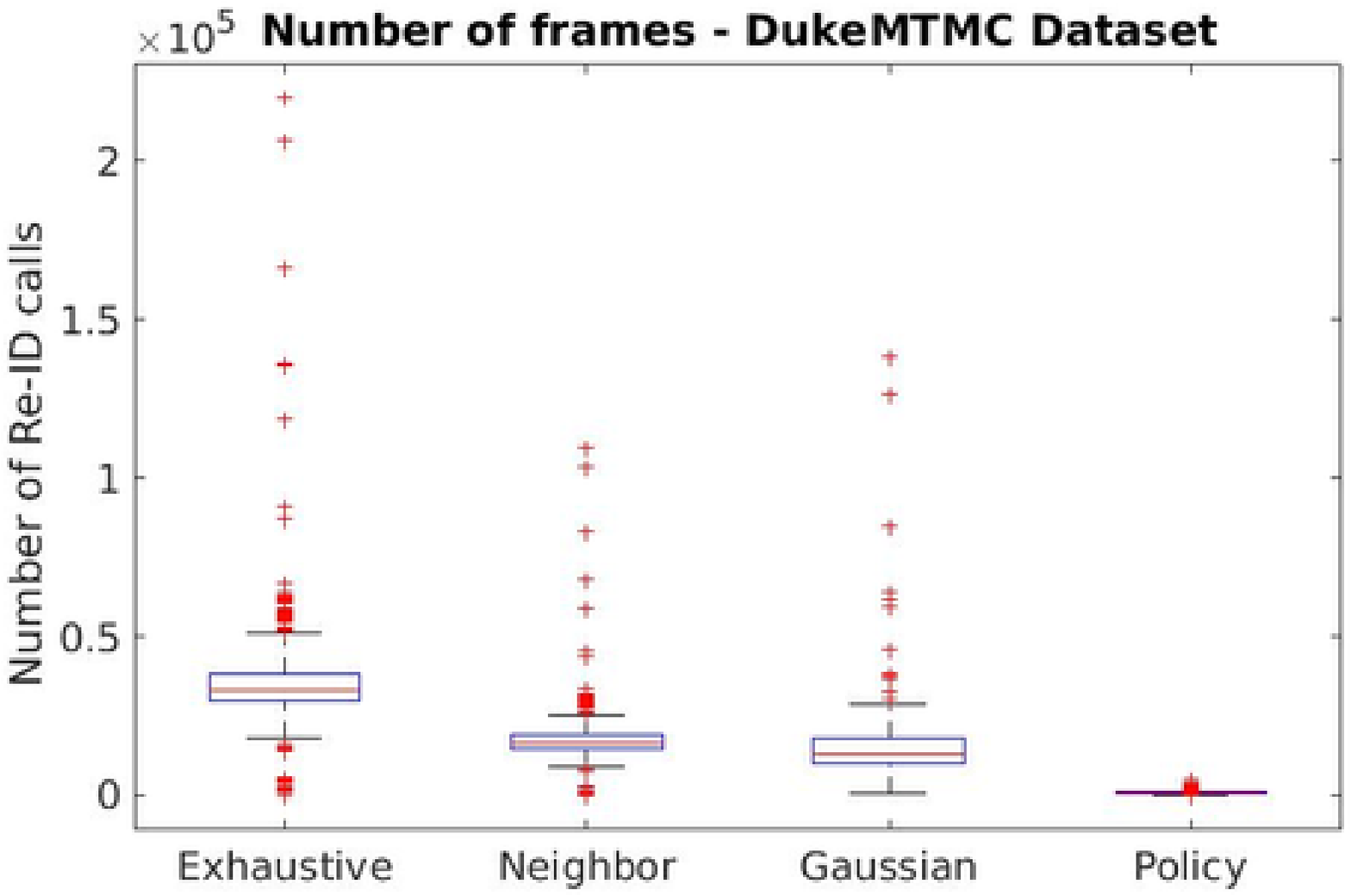}
\caption{Number of frames polled (F, equation~\ref{eq:f}) on DukeMTMC dataset for our deep RL based policy and its comparison with other baseline approaches.  }
\label{fig:f_metric}
\end{figure*}

Figure~\ref{fig:tran_traj_db3} show the sequence of cameras polled by the policy as compared to what is seen in the ground truth. Horizontal axis is time and vertical axis shows the camera schedules in ground truth ($GT$) and polled by policy ($Sel$). The dark colors are camera schedules (mapped with colormap) and white color shows the length of the transition. The figure reflects the performance of deep RL policy for making camera selection decisions. 
One important aspect of target tracking in multiple cameras is computational time. Many related methods match target template across neighboring cameras~\cite{SGM,nlpr_crf}, all cameras~\cite{nlpr_usc,deepcc} for offline tracking. However, such approach will require a large amount of frames to be processed for template matching. Using the proposed policy, this template matching will be limited to a single camera per time-step per person. In figure~\ref{fig:f_metric}, we have compared the number of frames to be processed of various such approaches. The figure shows the boxplot of $F$-metric scores computed over all targets using the deep RL policy and various baseline approaches on DukeMTMC dataset. 
$8$ cameras). 

\subsection{Impact of Camera Selection Decisions on Target Tracking in Camera Networks}

Now we will show the effectiveness of the camera selection decisions to enable target tracking in a camera network. To complete the tracking pipeline, we simulate the presence block of our proposed architecture. To simulate the presence block errors in a typical re-identification pipeline are generated by wrongly identifying the target with other available objects. We will compare the performance with state-of-the-art tracking methods. 

To perform this experiment, we have initialized the state vector with the initial location of the target and history vector being all zeros. The learned policy then polls a camera frame which is looked for the presence of the target using presence block (refer to section~\ref{sec:method_arch}). Unlike previous experiment, we are simulating a real re-identification pipeline for the presence block by adding errors to the presence decision. For example, to simulate $x\%$ error in re-identification, with probability $x$, we are taking another target's bounding box otherwise we are using the correct bounding box of the target. Once the presence is identified, the state vector is updated using the state-transition function. The updated state vector is then used by the policy to poll another camera and the process repeats till the end of the target's trajectory or the end of the sequence. The predicted trajectory is the sequence of (c,b) i.e., camera and bounding box values. The predicted trajectory of the target is then used to compute the MCTA metric scores. We have compared the performance of the policy with simulated re-identification errors with various state-of-the-art methods on the NLPR MCT dataset. The MCTA scores are shown in the tables~\ref{tab:MCTA_ict}~and~\ref{tab:MCTA_sct_ict} for ICT alone case and SCT+ICT case respectively. In table~\ref{tab:MCTA_ict}, we have shown MCTA values for inter-camera tracking (ICT) only where the single-camera trajectory of the target is taken from the ground-truth. In table~\ref{tab:MCTA_sct_ict}, shows the overall performance of the various methods i.e., during both single-camera tracking (SCT) and inter-camera tracking (ICT). The same experiments are reported by the related methods on NLPR dataset. In comparison to other methods, our approach performs better in most cases at $0\%$ error in re-identification. For higher errors, our method (especially deep RL) starts performing worse than others. 
Also, the related approaches are multi-target and multi-camera (MTMC) tracking approach whereas ours is single-target and multi-camera tracking. Therefore, to make a fair comparison, we have extended our approach to MTMC as explained in section~\ref{sec:data}. Similarly, results for DukeMTMC dataset are shown in the table~\ref{tab:MCTA_duke}.

\begin{table}[h!]
\caption{The table is showing average MCTA values for both SCT+ICT and ICT alone case on the DukeMTMC dataset. OM signifies Out-of-Memory error. There are no related approaches that define the tracking performance on DukeMTMC dataset using MCTA scores.}
\label{tab:MCTA_duke}
\centering
 \begin{tabular}{|c| c c |}
 \hline
  Approach & ICT alone & SCT + ICT  
  \\ [0.5ex] 
 \hline\hline
 Exact RL & OM & OM  \\ 
 Deep RL-0 & 0.8027 & 0.8191 \\ 
 Deep RL-5 & 0.6438  & 0.6215 \\ 
 Deep RL-10 & 0.6140 & 0.5417 \\ 
 Deep RL-15 & 0.5879 & 0.4768 \\ 
 Deep RL-20 & 0.5493 & 0.4357 \\ 
 \hline
 \end{tabular}
\end{table}

\section{Discussion}\label{sec:discuss}
We have proposed an approach for intelligent camera selection for dealing with target handovers in multi-camera target tracking. Our initial work used exact RL methods ~\cite{sharmaICAPS} and extended it to approximate methods using Deep RL in order to deal with larger camera networks. 
The deep RL implementation make better camera selection decisions and can be used with larger camera networks. However, there are a few limitations of the proposed deep RL approach. First, the performance of deep RL approach is sensitive to errors in Re-id. 
This requires investigations in training the deep learning based policy with a real re-identification so that the policy can learn how to handle errors during tracking. 
Second, large transition times results in a policy that has heavily imbalanced action distributions, e.g., $C_x$ becoming the most frequent action. 
Hence, efforts should be applied in exploring methods to handle imbalanced action space. Third, the indefinite transition time of a target makes exploration difficult in deciding whether the target goes out of the camera network or will appear again. There is a scope of improvement in identifying such cases. 

\section{Conclusion}\label{sec:conclusion}
We highlighted that re-identification  queries in target tracking across camera networks can become a performance and computational bottleneck for practical systems. We proposed a solution that intelligently makes these queries by selecting cameras that are more likely to contain the target at a given time. We proposed a reinforcement learning based approach that learns a policy for camera selection based on previous actions and target location. We empirically show on two benchmark datasets that the proposed approach substantially reduces the number of frames queried, with negligible loss of tracking performance. 

\section{Acknowledgement}
We acknowledge Infosys Center for Artificial Intelligence (CAI) at IIIT-Delhi for its partial support for conducting this research work.

\bibliography{paper}

\end{document}